\documentclass[acmtog]{acmart}

\AtBeginDocument{%
  }

\copyrightyear{2026}
\acmYear{2026}
\setcopyright{cc}
\setcctype{by-nc-nd}
\acmConference[SA Workshops '26]{SIGGRAPH Asia 2026 Workshops}{December 01--04, 2026}{Kuala Lumpur, Malaysia}
\acmBooktitle{SIGGRAPH Asia 2026 Workshops (SA Workshops '26), December 01--04, 2026, Kuala Lumpur, Malaysia}
\acmDOI{10.1145/3829341.3850009}
\acmISBN{979-8-4007-2843-3/2026/12}

\usepackage{colortbl}
\usepackage{pifont}
\colorlet{best}{orange!30}
\colorlet{second}{orange!12}

\begin{document}

\title{Sparse-View 4D Gaussian Splatting via Spatiotemporal Priors and
  Generative Assistance}

\author{Shengqi Wang}
\orcid{0009-0004-2786-4623}
\email{shengqi-21@mails.tsinghua.edu.cn}
\affiliation{%
  \department{Department of Automation}
  \institution{Tsinghua University}
  \city{Beijing}
  \country{China}
}

\author{Zhengxian Yang}
\email{zx-yang23@mails.tsinghua.edu.cn}
\affiliation{%
  \department{Department of Automation}
  \institution{Tsinghua University}
  \city{Beijing}
  \country{China}
}

\author{Kaiwen Tian}
\orcid{0009-0005-8397-1754}
\email{tkw23@mails.tsinghua.edu.cn}
\affiliation{%
  \department{Department of Automation}
  \institution{Tsinghua University}
  \city{Beijing}
  \country{China}
}

\author{Yang Liu}
\email{liuyang1605@jd.com}
\affiliation{%
  \institution{JD.com}
  \city{Beijing}
  \country{China}
}

\author{Bowen Liu}
\email{bowenliu@lusterinc.com}
\affiliation{%
  \institution{Beijing MEET YUAN Co., Ltd}
  \city{Beijing}
  \country{China}
}

\author{Hua Du}
\email{HUADU@LUSTERINC.COM}
\affiliation{%
  \institution{Beijing MEET YUAN Co., Ltd}
  \city{Beijing}
  \country{China}
}

\author{Taicheng Huang}
\email{huangtaicheng.1@jd.com}
\affiliation{%
  \institution{JD.com}
  \city{Beijing}
  \country{China}
}

\author{Jiamin Wu}
\orcid{0000-0003-3479-1026}
\email{wujiamin@tsinghua.edu.cn}
\affiliation{%
  \institution{Tsinghua University}
  \city{Beijing}
  \country{China}
}

\author{Tao Yu}
\correspondingauthor
\orcid{0000-0002-3818-5069}
\email{ytrock@mail.tsinghua.edu.cn}
\affiliation{%
  \institution{Tsinghua University}
  \city{Beijing}
  \country{China}
}

\renewcommand{\shortauthors}{Wang et al.}

\begin{abstract}
We present a 4D Gaussian Splatting framework for the Sparse-View Track of the \textit{SIGGRAPH Asia 2026 Volumetric Video Challenge}, which requires dynamic scene reconstruction from only six cameras with wide baselines. To achieve robust dynamic reconstruction under such sparse views, our framework integrates three components. \textbf{(1) Region-adaptive spatial priors:} We use foreground masks to guide Gaussian initialization and mask voting to control densification separately for the dynamic foreground and static background. Background geometry is regularized using monocular depth aligned to metric scale. \textbf{(2) Motion-consistent temporal priors:} We provide supervision at intermediate times through frame interpolation and constrain projected Gaussian motion with estimated optical flow. \textbf{(3) Generative assistance:} We place virtual cameras in the widest angular gaps and restore their rendered images using a diffusion-based model conditioned on camera poses. The restored images are iteratively incorporated into training as pseudo-supervision. 
On the validation set, our framework improves full-frame PSNR from 25.60 dB for the baseline to 29.75 dB. 
On the official test benchmark, it achieves \textbf{30.04 dB} full-frame PSNR and \textbf{27.88 dB} foreground PSNR, ranking first overall in the Sparse-View Track.
\end{abstract}



\maketitle

\section{Introduction}
\label{sec:intro}

Dynamic 4D Gaussian Splatting (4DGS)~\cite{4dgs,wang2025freetimegs,yang2026imvid} enables real-time novel view synthesis of dynamic scenes, but typically assumes dense multi-view coverage. The Sparse-View Track of the \textit{SIGGRAPH Asia 2026 Volumetric Video Challenge} challenges this reliance by requiring dynamic reconstruction from only six wide-baseline cameras.

Under such sparse-view settings, 4DGS reconstruction faces three interrelated challenges.
~ \textbf{\textit{(1) Imbalanced capacity allocation}}: Uniform densification can overpopulate the background, which occupies most image pixels, while leaving the dynamic foreground underrepresented. This imbalance can lead to background overfitting and degrade foreground rendering at novel viewpoints.
~ \textbf{\textit{(2) Under-constrained motion}}: Rapid motion can produce blur and large inter-frame displacements, while residual timing differences across cameras cause nominally simultaneous images to capture different motion states. These factors limit the reliability of photometric supervision for continuous motion reconstruction.
~ \textbf{\textit{(3) Incomplete view coverage}}: Wide angular gaps and occlusions leave some regions weakly observed or unseen. Holes and incomplete structures may therefore persist despite improved capacity allocation and motion regularization.

\begin{figure*}[t]
  \centering
  \includegraphics[width=\textwidth]{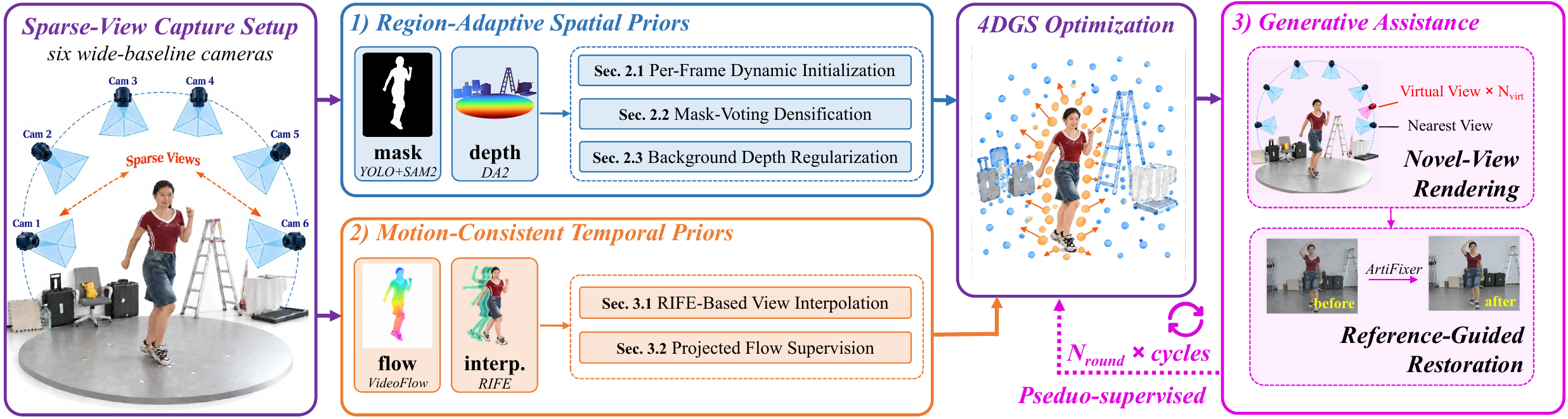}
  \caption{Overview of our sparse-view 4D Gaussian Splatting pipeline. Given videos from six cameras with wide baselines, our framework combines three components:
  (1)~\textit{region-adaptive spatial priors} use YOLO11 and SAM~2 masks for per-frame foreground initialization and mask voting during densification, and Depth Anything V2 predictions for background depth regularization;
  (2)~\textit{motion-consistent temporal priors} use RIFE interpolation to provide supervision at intermediate times and VideoFlow estimates to constrain projected Gaussian motion; and
  (3)~\textit{generative assistance} restores images rendered from virtual cameras in wide angular gaps using the ArtiFixer diffusion model~\cite{delutio2026artifixer}, with the nearest training views in camera pose space as references.
  The restored images are iteratively incorporated into training as pseudo-supervision, yielding the final 4DGS model for novel view synthesis.}
  \Description{Pipeline from six wide-baseline cameras to 4D Gaussian Splatting optimization. Spatial priors use foreground masks and monocular depth, temporal priors use interpolation and optical flow, and generative assistance restores virtual views before pseudo-supervised re-injection.}
  \label{fig:pipeline}
\end{figure*}

To address these interrelated challenges, we combine three complementary components (Fig.~\ref{fig:pipeline}):
~ \textbf{\textit{(1) Region-adaptive spatial priors}} (Sec.~\ref{sec:spatial}) use foreground masks to guide initialization, mask voting to adapt densification, and monocular depth aligned to metric scale to regularize background geometry.
~ \textbf{\textit{(2) Motion-consistent temporal priors}} (Sec.~\ref{sec:temporal}) provide supervision between captured frames through interpolation and constrain projected Gaussian velocities with estimated optical flow.
~ \textbf{\textit{(3) Generative assistance}} (Sec.~\ref{sec:gen}) places virtual cameras in the widest angular gaps, restores their rendered images using ArtiFixer~\cite{delutio2026artifixer}, and iteratively incorporates the restored views into training as pseudo-supervision.

Extensive validation confirms that our framework effectively stabilizes sparse dynamic reconstruction, ultimately earning \textbf{\#1} place in the official Sparse-View Track.

\section{Region-Adaptive Spatial Priors}
\label{sec:spatial}

We build on spatiotemporal Gaussians~\cite{yang2026imvid}, where each primitive has a temporal center $t_i$, a temporal extent $\tau_i$, and a 3D velocity $v_i$. To address imbalanced capacity allocation, we introduce region-adaptive priors that support detailed foreground reconstruction while limiting excessive densification and overfitting in the static background.

We precompute two scene priors: per-frame foreground masks $M_{\mathrm{fg}}$ obtained using YOLO11~\cite{jocher2023ultralytics} and tracked with SAM~2~\cite{ravi2024sam2}, and monocular depth predictions $D_{\mathrm{pred}}$ from DA2~\cite{yang2024depthanythingv2}. The masks guide Gaussian initialization and densification separately for the foreground and background, while the depth predictions regularize background geometry.

\subsection{Per-Frame Dynamic Initialization}
\label{sec:init}
Following~\cite{wang2025freetimegs}, we initialize Gaussian primitives by triangulating multi-view correspondences obtained with RoMa~\cite{edstedt2024roma}. To accommodate occlusions and segmentation noise, a triangulated point $p$ at time $t$ is labeled as foreground if its projection falls outside the foreground mask in at most $V_{\mathrm{tol}}$ views:
\begin{equation}
  \ell(p,t) =
  \mathbb{1}\left[
    \sum_{v \in \mathcal{V}}
    \left(1-M_{\mathrm{fg},v,t}(\pi_v(p))\right)
    \le V_{\mathrm{tol}}
  \right],
  \label{eq:mask-label}
\end{equation}
where $\mathcal{V}$ is the set of camera views used for voting, $\pi_v(p)$ denotes the 2D projection of $p$ into view $v$, and $M_{\mathrm{fg},v,t}$ is the corresponding binary foreground mask.

Foreground primitives are initialized at every frame, with temporal centers $t_i=t$ and short temporal extents $\tau_{\mathrm{fg}}$ to capture localized motion. In contrast, background primitives are initialized every $K_{\mathrm{bg}}$ frames with larger temporal extents $\tau_{\mathrm{bg}}$. Parameter values are provided in Sec.~\ref{sec:impl}.

\subsection{Mask-Voting Densification}
\label{sec:densify}
During training, we update each primitive's foreground or background label through multi-view mask voting, using the tolerance $V_{\mathrm{tol}}$ introduced in Sec.~\ref{sec:init}. Following the adaptive density control of 3DGS~\cite{kerbl3Dgaussians}, we densify primitives whose accumulated absolute view-space gradients~\cite{ye2024absgs} exceed the threshold assigned to their region. We use a lower threshold $\theta_{\mathrm{fg}}$ to encourage foreground refinement and a higher threshold $\theta_{\mathrm{bg}}$ to limit excessive background densification.

\subsection{Background Depth Regularization}
\label{sec:depth}
We align monocular depth predictions to metric scale using depth samples derived from triangulated RoMa points. RANSAC is used to estimate the scale $s$ and shift $b$, yielding $\hat{D} = s D_{\mathrm{pred}} + b$. To reduce the influence of potentially noisy foreground predictions, we restrict depth supervision to background pixels $\mathcal{B}$:
\begin{equation}
  \mathcal{L}_{\mathrm{depth}} =
  \frac{1}{|\mathcal{B}|}
  \sum_{u \in \mathcal{B}}
  \left|D_{\mathrm{rend}}(u)-\hat{D}(u)\right|,
  \label{eq:depth-loss}
\end{equation}
where $D_{\mathrm{rend}}$ denotes the rendered depth map. This loss is weighted by $\lambda_{\mathrm{depth}}$ (Sec.~\ref{sec:impl}).

\section{Motion-Consistent Temporal Priors}
\label{sec:temporal}

While spatial priors improve capacity allocation, photometric supervision alone may still yield ambiguous Gaussian motion estimates, particularly in the presence of rapid motion and residual timing differences across cameras. We introduce two complementary temporal priors, both computed offline. Intermediate frames generated by RIFE~\cite{Huang2022RIFE} provide photometric supervision between captured frames, while optical flow estimated by VideoFlow (BOFNet)~\cite{shi2023videoflow} regularizes projected Gaussian motion.

\subsection{RIFE-Based View Interpolation}
\label{sec:rife}
For each pair of consecutive captured frames at times $t_j$ and $t_{j+1}$, we use RIFE to generate $K_{\mathrm{rife}}-1$ intermediate frames at times $t_{j,r}=t_j+\frac{r}{K_{\mathrm{rife}}}(t_{j+1}-t_j)$, where $r=1,\ldots,K_{\mathrm{rife}}-1$. These images provide photometric pseudo-supervision between captured frames.

During training, each epoch visits every camera--time-slot pair $(c,j)$ once and selects an interpolation phase:
\begin{equation}
  r^\star = (j+e+c) \bmod K_{\mathrm{rife}},
  \label{eq:phase}
\end{equation}
where $e$ and $c$ are the epoch and camera indices, respectively. The phase $r^\star=0$ selects the captured frame at $t_j$, while the remaining phases select interpolated frames. This schedule cycles through the phases across epochs. During the final $E_{\mathrm{rife}}$ epochs, we use only captured frames to reduce the influence of interpolation artifacts.

\subsection{Projected Flow Supervision}
\label{sec:flowsup}
The instantaneous 3D velocity $v_i$ of each Gaussian is mapped onto the 2D image plane via the perspective projection Jacobian~\cite{yang2026imvid}:
\begin{equation}
  f_i^{2D} = \mathbf{J} (\mathbf{R} v_i), \quad
  \mathbf{J} = \begin{bmatrix}
    f_x / Z & 0 & -f_x X / Z^2 \\[2pt]
    0 & f_y / Z & -f_y Y / Z^2
  \end{bmatrix},
  \label{eq:project-flow}
\end{equation}
where $\mathbf{R}$ is the world-to-camera rotation, $(X, Y, Z)$ denotes the Gaussian center in camera coordinates and $(f_x, f_y)$ are the focal lengths.
Differentiable alpha compositing yields the rendered 2D optical flow field, $\mathcal{F}_{\mathrm{rend}} = \sum_i f_i^{2D} \alpha_i \prod_{j<i} (1 - \alpha_j)$.
We supervise $\mathcal{F}_{\mathrm{rend}}$ by minimizing the endpoint error (EPE) against pseudo-ground-truth flow fields $\mathcal{F}_{\mathrm{pred}}$ estimated by VideoFlow:
\begin{equation}
  \mathcal{L}_{\mathrm{flow}} = \|\mathcal{F}_{\mathrm{rend}} -
  \mathcal{F}_{\mathrm{pred}}\|_2,
  \label{eq:flow-loss}
\end{equation}
weighted by $\lambda_{\mathrm{flow}}$ (Sec.~\ref{sec:impl}).

\section{Generative Assistance}
\label{sec:gen}

Despite spatial and temporal regularization, \textit{incomplete view coverage} due to wide angular gaps and occlusions can still lead to missing structures and floaters at novel viewpoints. To provide additional supervision at these viewpoints, we render images from virtual cameras and restore them using a diffusion model conditioned on camera poses and reference views. The restored images are iteratively incorporated into 4DGS training as pseudo-supervision.

\subsection{Reference-Guided Completion}
\label{sec:restore}

We establish $N_{\mathrm{virt}}$ virtual cameras by recursively bisecting the widest angular gaps along the inward-looking arc, yielding intermediate viewpoints with known camera poses.

For each virtual view $v$, we select the nearest training views $\mathcal{R}(v)$ according to the pose distance
\begin{equation}
  d(v,r) =
  \|C_v-C_r\|_2 +
  \lambda_{\mathrm{rot}}\angle(R_v,R_r),
  \label{eq:ref-select}
\end{equation}
where $C$ and $R$ denote camera centers and orientations, respectively, $\angle(\cdot,\cdot)$ measures the relative rotation angle, and $\lambda_{\mathrm{rot}}$ balances positional and rotational differences.

We then use ArtiFixer~\cite{delutio2026artifixer}, denoted by $\mathcal{D}$, to restore the rendered image $I_v^{\mathrm{rend}}$ using the selected reference images. The resulting image
$\hat{I}_v = \mathcal{G}(I_v^{\mathrm{rend}}, \{I_r\}_{r \in \mathcal{R}(v)})$
serves as a pseudo-target for subsequent 4DGS optimization.

\subsection{Iterative Re-injection}
\label{sec:reinject}
The restored virtual views are incorporated into training under the standard photometric and depth supervision, providing additional pseudo-supervision at under-observed viewpoints.
Specifically, reconstruction and generative restoration operate in a bootstrapping loop for $N_{\mathrm{round}}$ cycles: once a model is reconstructed, novel-view renders from the virtual cameras are restored by $\mathcal{D}$, and a new 4DGS model is trained using both captured views and the restored pseudo-targets. As the representation improves across rounds, cleaner novel-view renders provide higher-quality conditioning for the diffusion model, yielding progressively superior pseudo supervision.
To mitigate potential generative hallucinations and promote fidelity to ground-truth observations, during the last $E_{\mathrm{tail}}$ epochs of each training cycle, the model is fine-tuned on authentic captured views only. Hyperparameter values are provided in Sec.~\ref{sec:impl}.

\section{Implementation Details}
\label{sec:impl}

We train for 30 epochs at half resolution and evaluate at full resolution. Each epoch visits every camera--time-slot pair once, following the sampling schedule in Sec.~\ref{sec:rife}. We use spherical harmonics (SH) of degree $1$ and set the near clipping distance to $0.5$.
For region-adaptive spatial priors, we set
$V_{\mathrm{tol}} = 3$, $K_{\mathrm{bg}} = 10$, initial $\tau_{\mathrm{fg}} / \tau_{\mathrm{bg}} = 0.01\mathrm{s} / 1.0\mathrm{s}$,
$\theta_{\mathrm{fg}} = 5\times 10^{-6}$, $\theta_{\mathrm{bg}} =
1\times 10^{-5}$, and $\lambda_{\mathrm{depth}} = 1.0$.
For motion-consistent temporal priors, we set $K_{\mathrm{rife}} = 16$,
$E_{\mathrm{rife}} = 2$, and $\lambda_{\mathrm{flow}} = 1.0$.
For generative assistance, we use $N_{\mathrm{virt}} = 14$,
$N_{\mathrm{round}} = 2$, and $E_{\mathrm{tail}} = 2$.

\section{Experiments and Analysis}
\label{sec:exp}

\begin{table}[t]
  \centering
  \caption{Ablation study on the validation set.
  ``Baseline'' refers to the reconstruction method in~\cite{yang2026imvid}.
  ``Ours (recon)'' includes all proposed spatial and temporal priors,
  while ``Ours (+ gen)'' additionally uses generative assistance.
  Parenthesized $\theta$ indicates a uniform densification threshold;
  otherwise, $\theta_{\mathrm{fg}}$ and $\theta_{\mathrm{bg}}$ are used
  for the foreground and background, respectively.
  {\setlength{\fboxsep}{1pt}\colorbox{best}{Best} and
  \colorbox{second}{second-best}} results are highlighted for each metric.}
  \label{tab:main}

  \setlength{\tabcolsep}{2.5pt}
  \small
  \begin{tabular}{@{}lcccccc@{}}
    \toprule
    & \multicolumn{3}{c}{Full frame}
    & \multicolumn{3}{c}{Foreground} \\
    \cmidrule(lr){2-4} \cmidrule(lr){5-7}
    Setting
    & PSNR$\uparrow$ & SSIM$\uparrow$ & LPIPS$\downarrow$
    & PSNR$\uparrow$ & SSIM$\uparrow$ & LPIPS$\downarrow$ \\
    \midrule

    Baseline ($\theta{=}5\mathrm{e}{-6}$)
    & 25.60 & 0.9091 & 0.2429
    & 27.28 & 0.8686 & 0.2276 \\

    Baseline ($\theta{=}1\mathrm{e}{-5}$)
    & 28.08 & 0.9214 & 0.2148
    & 26.63 & 0.8554 & 0.2539 \\

    \midrule

    w/o voting ($\theta{=}5\mathrm{e}{-6}$)
    & 28.27 & 0.9263 & 0.1977
    & 27.68 & 0.8741 & 0.2220 \\

    w/o voting ($\theta{=}1\mathrm{e}{-5}$)
    & 28.49 & 0.9259 & 0.2017
    & 26.92 & 0.8614 & 0.2409 \\

    w/o mask init
    & \cellcolor{second}28.64
    & \cellcolor{second}0.9267
    & 0.1955
    & 27.68 & 0.8743 & 0.2219 \\

    w/o depth
    & 26.52 & 0.9168 & 0.2274
    & 27.48 & 0.8714 & 0.2252 \\

    w/o flow
    & 28.46 & 0.9266 & 0.1997
    & 27.64 & 0.8739 & 0.2213 \\

    w/o RIFE
    & 28.49 & 0.9266 & 0.1994
    & 27.68 & 0.8735 & 0.2210 \\

    \midrule

    Ours (recon)
    & 28.63
    & \cellcolor{best}\textbf{0.9268}
    & \cellcolor{second}0.1938
    & \cellcolor{second}27.90
    & \cellcolor{best}\textbf{0.8764}
    & \cellcolor{second}0.2201 \\

    Ours (+ gen)
    & \cellcolor{best}\textbf{29.75}
    & 0.9248
    & \cellcolor{best}\textbf{0.1869}
    & \cellcolor{best}\textbf{27.96}
    & \cellcolor{second}0.8745
    & \cellcolor{best}\textbf{0.2163} \\

    \bottomrule
  \end{tabular}
\end{table}

\paragraph{Experimental Setup}
\label{sec:setup}
We evaluate on the official validation set, using six calibrated cameras for training and eight held-out cameras for novel view evaluation. Following the challenge protocol, we report PSNR, SSIM, and LPIPS with an AlexNet backbone on both full frames and isolated dynamic foreground regions. All 4DGS models are trained on a single NVIDIA RTX 3090 GPU.

\begin{figure}[t]
  \centering
  \begin{minipage}[b]{0.315\linewidth}
    \centering
    \includegraphics[width=0.92\linewidth]{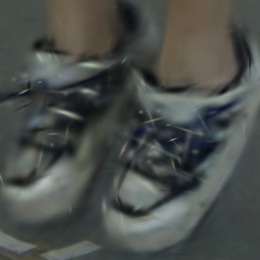}\\[-4pt]
    {\footnotesize (a) w/o voting ($\theta{=}1\mathrm{e}{-5}$)}
  \end{minipage}
  \hfill
  \begin{minipage}[b]{0.315\linewidth}
    \centering
    \includegraphics[width=0.92\linewidth]{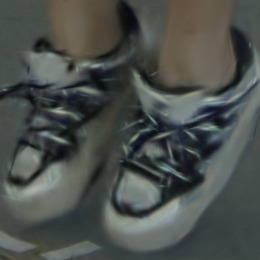}\\[-4pt]
    {\footnotesize (b) Ours (recon)}
  \end{minipage}
  \hfill
  \begin{minipage}[b]{0.315\linewidth}
    \centering
    \includegraphics[width=0.92\linewidth]{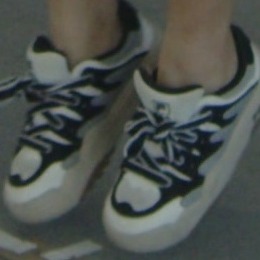}\\[-4pt]
    {\footnotesize (c) Ground Truth}
  \end{minipage}
  \vspace{-5pt}
  \caption{Qualitative ablation of mask-voting densification.}
  \Description{Qualitative comparison of dynamic foreground reconstruction without mask voting versus our full adaptive model.}
  \label{fig:spatial}
\end{figure}

\begin{figure}[t]
  \centering
  \begin{minipage}[b]{0.315\linewidth}
    \centering
    \includegraphics[width=0.92\linewidth]{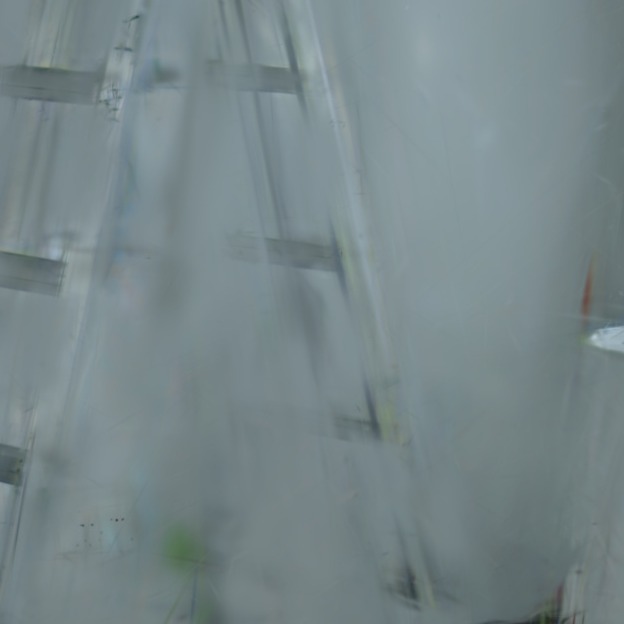}\\[-4pt]
    {\footnotesize (a) w/o depth}
  \end{minipage}
  \hfill
  \begin{minipage}[b]{0.315\linewidth}
    \centering
    \includegraphics[width=0.92\linewidth]{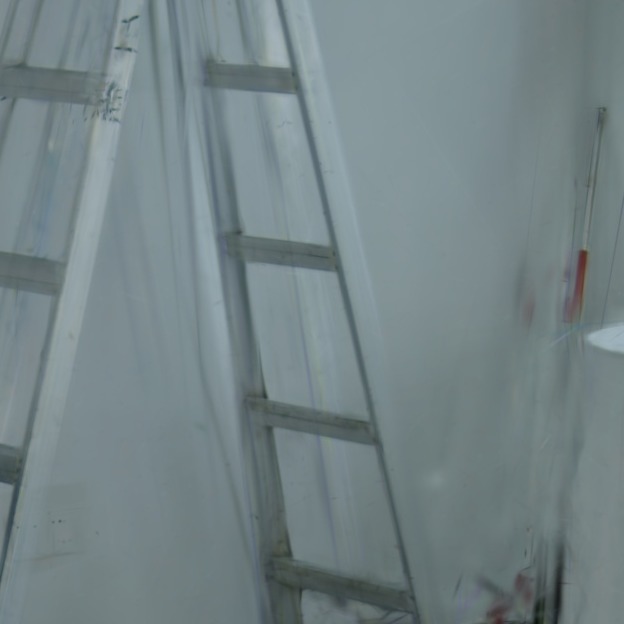}\\[-4pt]
    {\footnotesize (b) Ours (recon)}
  \end{minipage}
  \hfill
  \begin{minipage}[b]{0.315\linewidth}
    \centering
    \includegraphics[width=0.92\linewidth]{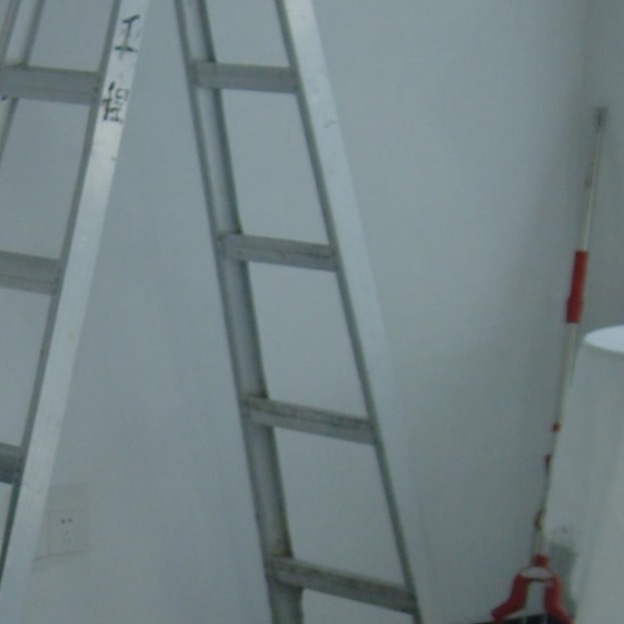}\\[-4pt]
    {\footnotesize (c) Ground Truth}
  \end{minipage}
  \vspace{-5pt}
  \caption{Qualitative ablation of background depth regularization.}
  \Description{Qualitative comparison of background reconstruction without depth regularization versus our full model.}
  \label{fig:depth}
\end{figure}

\paragraph{Addressing Region-Dependent Requirements}
The two \textit{w/o voting} variants in Table~\ref{tab:main} illustrate a trade-off between full-frame and foreground quality under a uniform densification threshold. Lowering $\theta$ from $1\times10^{-5}$ to $5\times10^{-6}$ improves foreground PSNR from 26.92 to 27.68~dB but reduces full-frame PSNR from 28.49 to 28.27~dB. Mask-voting densification balances these regional requirements, achieving 28.63~dB full-frame PSNR and 27.90~dB foreground PSNR, with finer shoe details in Fig.~\ref{fig:spatial}.
Removing foreground-aware initialization (\textit{w/o mask init}) reduces foreground PSNR from 27.90 to 27.68~dB, while full-frame PSNR remains nearly unchanged as background pixels dominate. Removing background depth regularization (\textit{w/o depth}) reduces full-frame PSNR by 2.11~dB and produces more pronounced artifacts around ladder edges (Fig.~\ref{fig:depth}). Note that, compared with the conservative baseline ($\theta{=}1\mathrm{e}{-5}$), the remaining priors introduce additional motion and pseudo-supervision signals without directly anchoring background geometry, making the weakly observed background more susceptible to geometric drift and causing a lower full-frame score.

\begin{figure}[t]
  \centering
  \begin{minipage}[b]{0.315\linewidth}
    \centering
    \includegraphics[width=0.92\linewidth]{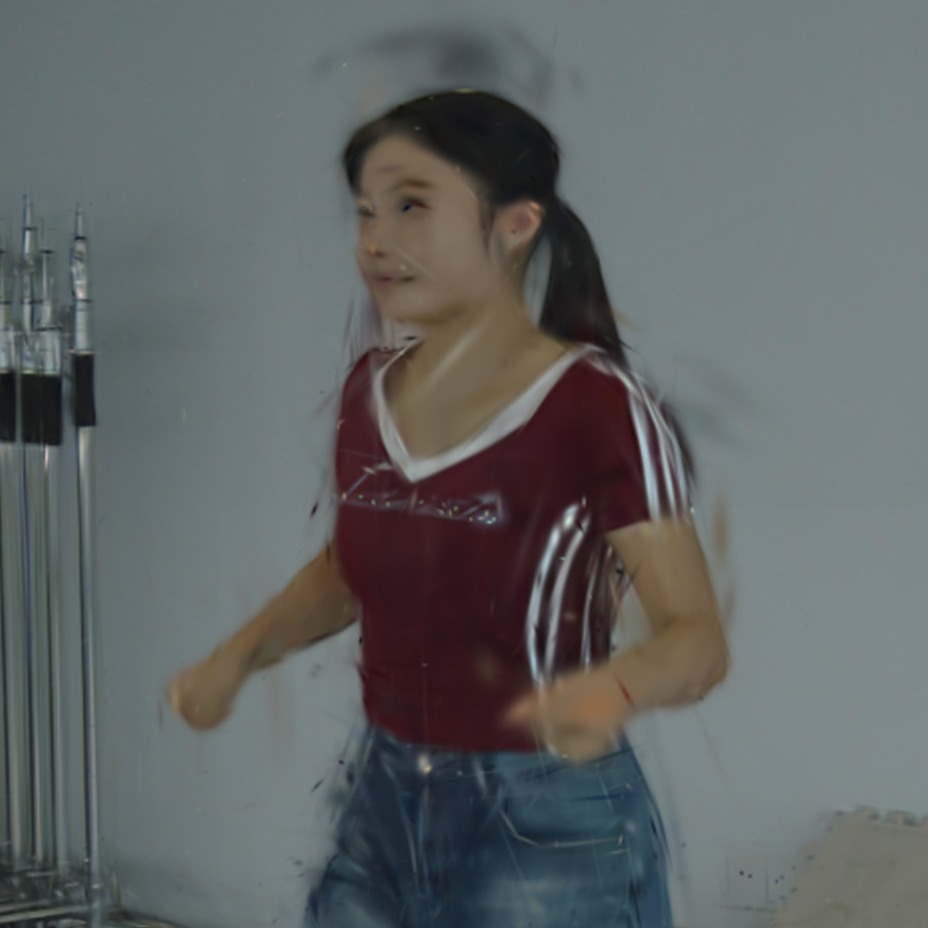}\\[-4pt]
    {\footnotesize (a) w/o temporal}
  \end{minipage}
  \hfill
  \begin{minipage}[b]{0.315\linewidth}
    \centering
    \includegraphics[width=0.92\linewidth]{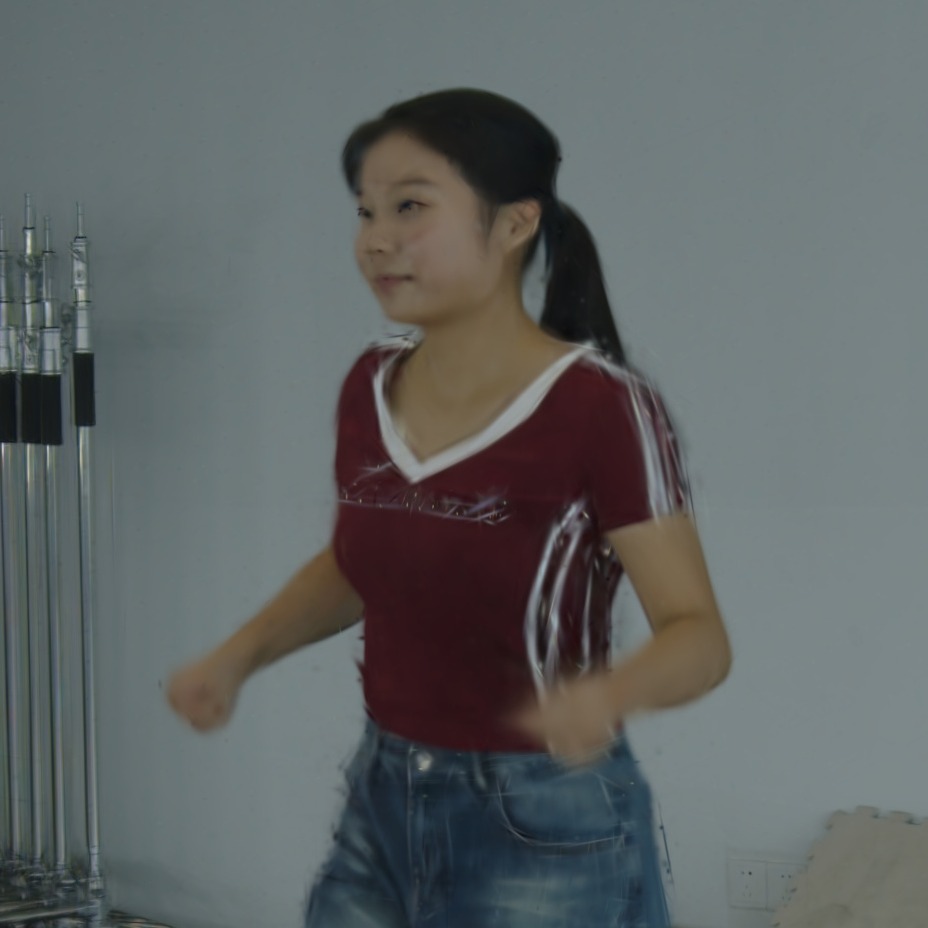}\\[-4pt]
    {\footnotesize (b) Ours (recon)}
  \end{minipage}
  \hfill
  \begin{minipage}[b]{0.315\linewidth}
    \centering
    \includegraphics[width=0.92\linewidth]{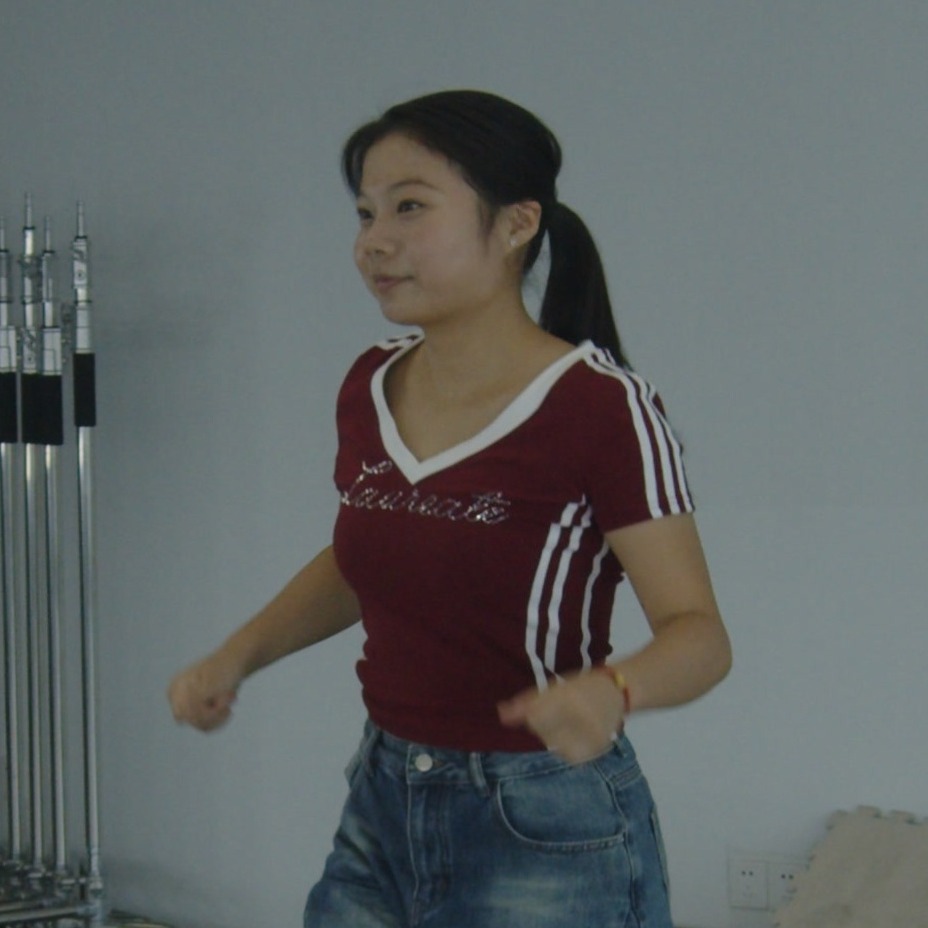}\\[-4pt]
    {\footnotesize (c) Ground Truth}
  \end{minipage}
  \vspace{-5pt}
  \caption{Qualitative ablation of motion-consistent temporal priors.}
  \Description{Qualitative comparison of dynamic reconstruction without temporal priors versus our full model.}
  \label{fig:temporal}
\end{figure}

\paragraph{Constraining Dynamic Motion}
As shown in Table~\ref{tab:main}, projected optical flow supervision and RIFE interpolation each contribute $\sim$0.2~dB across metrics. Without these temporal constraints, dynamic regions tend to suffer from motion blur and floaters. Incorporating our motion-consistent temporal priors yields cleaner object boundaries and suppresses floater artifacts. This qualitative improvement is especially noticeable in time segments featuring rapid motion, where inter-frame displacements are large (Fig.~\ref{fig:temporal}).

\begin{figure}[t]
  \centering
  \begin{minipage}[b]{0.315\linewidth}
    \centering
    \includegraphics[width=0.92\linewidth]{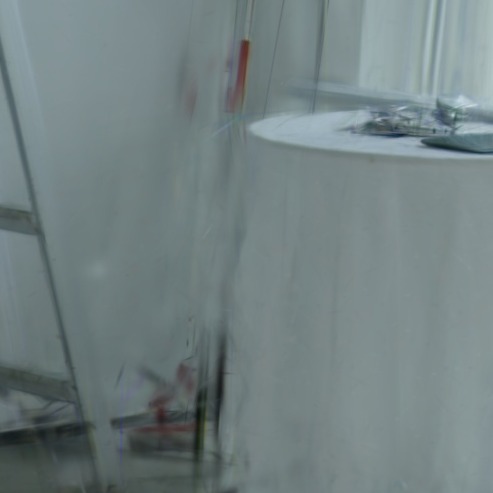}\\[-4pt]
    {\footnotesize (a) Ours (recon)}
  \end{minipage}
  \hfill
  \begin{minipage}[b]{0.315\linewidth}
    \centering
    \includegraphics[width=0.92\linewidth]{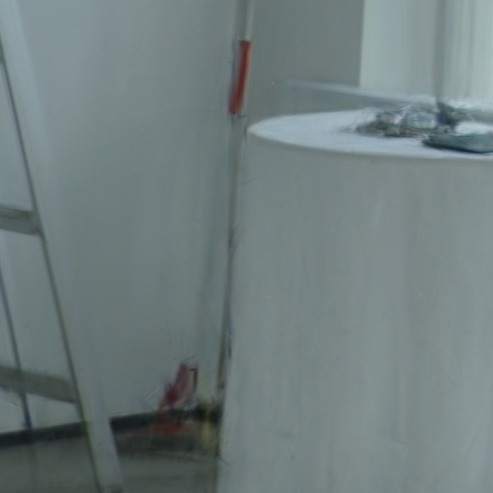}\\[-4pt]
    {\footnotesize (b) Ours (+ gen)}
  \end{minipage}
  \hfill
  \begin{minipage}[b]{0.315\linewidth}
    \centering
    \includegraphics[width=0.92\linewidth]{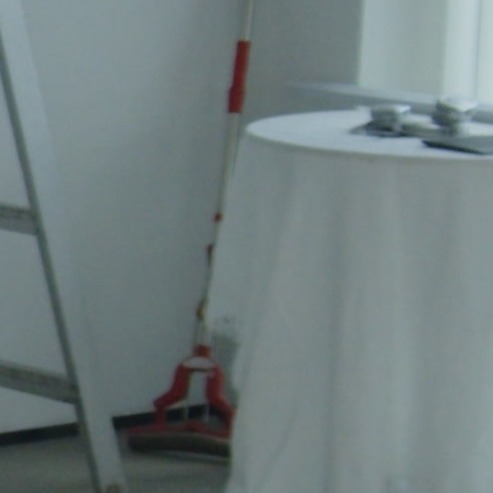}\\[-4pt]
    {\footnotesize (c) Ground Truth}
  \end{minipage}
  \vspace{-5pt}
  \caption{Qualitative ablation of generative assistance.}
  \Description{Qualitative comparison of novel-view reconstruction with and without generative assistance.}
  \label{fig:gen}
\end{figure}

\paragraph{Overcoming Incomplete View Coverage}
As shown in Table~\ref{tab:main} and Fig.~\ref{fig:gen}, iterative generative assistance successfully completes under-observed regions—such as the mop tucked in the corner occluded by the ladder and table—while substantially suppressing floaters, improving full-frame PSNR by +1.12~dB (to 29.75~dB) and improving perceptual LPIPS. Although SSIM slightly decreases due to subtle high-frequency variations introduced by diffusion completion, the overall visual coherence and fidelity improve notably.

\section{Conclusion and Future Work}
\label{sec:conclusion}

We presented a 4D Gaussian Splatting framework that combines region-adaptive spatial priors, motion-consistent temporal priors, and generative assistance to reconstruct dynamic scenes from six cameras with wide baselines. Our method achieves high-quality novel view synthesis and ranks first overall in the Sparse-View Track of the SIGGRAPH Asia 2026 Volumetric Video Challenge. Future work will focus on reducing reconstruction time and model size to support real-time applications on standalone VR headsets.

\bibliographystyle{ACM-Reference-Format}
\bibliography{refs}

\end{document}